# An Offline Technique for Localization of License Plates for Indian Commercial Vehicles

Satadal Saha [1], Subhadip Basu [2], Mita Nasipuri [2], Dipak Kumar Basu [#2]
[#] AICTE Emeritus Fellow
[1] CSE Department, MCKV Institute of Engineering, Howrah, India
[2] CSE Department, Jadavpur University, Kolkata, India

*Abstract*—**Automatic License Plate Recognition (ALPR) is a challenging area of research due to its importance to variety of commercial applications. The overall problem may be subdivided into two key modules, firstly, localization of license plates from vehicle images, and secondly, optical character recognition of extracted license plates. In the current work, we have concentrated on the first part of the problem, i.e., localization of license plate regions from Indian commercial vehicles as a significant step towards development of a complete ALPR system for Indian vehicles. The technique is based on color based segmentation of vehicle images and identification of potential license plate regions. True license plates are finally localized based on four spatial and horizontal contrast features. The technique successfully localizes the actual license plates in 73.4% images.**

## I. INTRODUCTION

*Automatic License Plate Recognition* from vehicle images has long been an active area for the researchers. In general, objective of such systems is to localize the license plate region(s) from the vehicle images, captured through a road-side camera, and interpret them using an *Optical Character Recognition* (OCR) system.

ALPR systems are widely implemented for automatic ticketing of vehicles at car parking facilities, tracking vehicles during traffic signal violations and related applications with huge saving of human energy and cost. Any ALPR system may be broadly categorized into two types, namely, an online ALPR system and an offline ALPR system. In an online ALPR system, the localization and interpretation of license plates take place instantaneously from the incoming video frames, enabling real-time tracking of moving vehicles through the surveillance camera. An offline ALPR system, in contrast, captures the vehicle images and stores them in a centralized data server for further processing, i.e. , for interpretation of vehicle license plates. The current work, discussed in this paper, comes under the later category of solutions.

Various techniques have been developed recently for the purpose for efficient detection of license plate regions from offline vehicular images. Most of these works [1-4] concentrate on localizing standardized license plate regions using edge based features. Some of these works [2, 5, 6] use the image of a vehicle, well placed in front of a camera, to get a clear view of the license plate. But in the practical scenario, there may be multiple vehicles of different types in a single scene along with partial occlusions of the license plates from other objects. In one of the earlier works [1], Rank filter is used for localization of license plate regions giving bad result for skewed license plates. An analysis of Swedish license plate is done in [2] using vertical edge detection followed by binarisation. This does not give better result for non-uniformly illuminated plates. An exhaustive study of plate recognition is done in [3] for different European countries. In Greece the license plate uses shining plate. The bright white background is used as a characteristic for license plate in [4]. Spanish license plate is recognized in [5] using Sobel edge detection operator. It also uses the aspect ratio and distance of the plate from the center of the image as characteristics. But it is constrained for single line license plates. During the localization phase the position of the characters is used in [6]. It assumes that no significant edge lies near the license plate and characters are disjoint.

In the developed countries and in most of the developing countries the attributes of the license plates are strictly maintained. For example, the size of the plate, color of the plate, font face / size / color of each character, spacing between subsequent characters, the number of lines in the license plate, script etc. are maintained very specifically. Some of the images of standard license plates, used in developed countries, are shown in Fig 1 (a). However, in India, the license plates are not yet standardized across different states, making localization and subsequent recognition of license plates extremely difficult. Moreover, in India license plates are often written in multiple scripts. Fig. 1(b) shows some of the typical Indian license plates with variations in shape, size, script etc. This large diversity in the features of the license plate makes its localization a challenging problem for the research community.

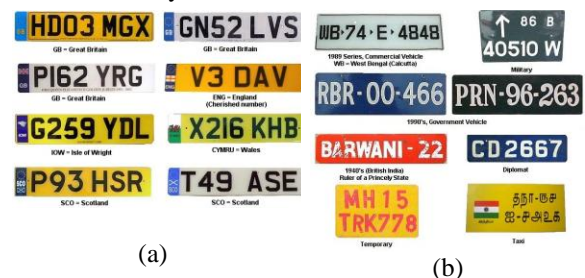

(a)  (b)

Fig. 1. License Plate images.

(a). Standardized license plates of European vehicles
(b). License plates of Indian vehicles

Two types of license plates are used in India. For commercial vehicles, the plate has a yellow background and black numbering. For private vehicles a white background with black numbering is used. The current Indian vehicle registration scheme comprises of a two-letter identification code for the state, in which the vehicle is registered. It is followed by a two-digit numeric code to identify the district. In the union territories and the erstwhile union territory of Delhi, the district code is omitted. This is often followed by a series code, e.g. 14C is the fourteenth series for private cars and 2M is the second series for motorbikes. Recently many states have been adapting the dual letter series code system, for example car series' are CA, CB, CC; motorbike series' are MA, MB and so on. Finally a four-digit number is used to uniquely identify the vehicle. Most states however still use the standard series code, denoted by a single letter of the alphabet. When the alphabet reaches Z, the length of the prefix is increased to 2. So after WB-02 9999, the next number is WB-02 A 0001 and after WB-02 Z 9999 it is WB-02 AA 0001 and so on.

Not much work has been done on detecting the license plates of Indian vehicles. A reference on ALPR of Indian vehicles is found in [7]. However, the technical details and the performance of such techniques are yet to be evaluated. In the light of above facts, the objective of the paper is to present a novel technique for localization of license plate regions, an important step towards development of a complete ALPR system, from Indian commercial vehicles. In the following section we discuss the basic methodologies employed for the image preprocessing, segmentation and the localization task. In the subsequent sections the experimental results and the conclusions are discussed.

## II. PRESENT WORK

In our present work we acquired the images using a digital camera placed by the road side facing towards the incoming vehicles so that the frontal image of vehicles can be obtained. The snaps are taken automatically at a regular interval of 1 second. This image capturing process is done at three different roads of Kolkata at different lighting conditions, even at night. The schematic flow chart of the developed technique, discussed in this paper, is shown in Fig. 2.

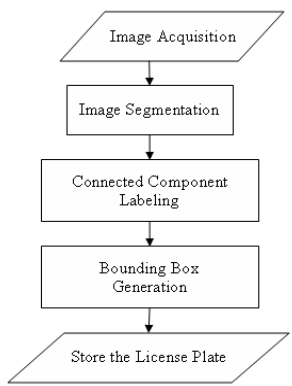

Fig. 2. A schematic work flow of the developed technique is shown.

Once all the images have been captured, the rest of the processing is done in an offline mode. We have first processed the image to make some refinement over it. RGB and HSI color models are used for all the processes done over the images. The [R, G, B] value of a pixel in the image is read from the file and the [H, S, I] value is calculated using the following conversion formula:

$$H = \begin{cases} \theta & if\ B \leq G \\ 360 - \theta & if\ B \leq G \end{cases}$$

$$\theta = \cos^{-1}\left[\frac{\frac{1}{2}[(R-G)+(R-B)]}{[(R-G)^2+(R-B)(G-B)]^{\frac{1}{2}}}\right]$$

$$S = 1 - \frac{3}{(R+G+B)}[\min(R,G,B)]$$

$$I = \frac{1}{3}(R+G+B)$$

(1)

where R, G, B, H, S and I are red color content, green color content, blue color content, hue, saturation and intensity of a pixel. Here it is assumed that R, G, and B are normalized within the range [0, 1]. Hue indicates true color of the pixel, saturation indicates the purity of the color and intensity indicates the brightness of the pixel. Hue is measured in terms of angle made by the color line with the red axis and varies from 0 to 360 degrees. The saturation varies from 0 to 1. Saturation equals to 0 means pure black and saturation equals to 1 means pure color as made by hue. The intensity also varies from 0 to 1. Intensity equals to 0 means pure black and intensity equals to 1 means pure white color. Obviously the license plates have a standard color thereby having a fixed hue and saturation with a little variation over them. The intensity component over the license plate varies highly over the edges of the character written on it. The main idea of our present work is to localize the license plate considering the hue and saturation and then calculate the contrast over the suspected regions using the intensity variation over them. Definitely the true license plate region will produce more contrast per unit area than the other places. This will identify the true license plate region over the image. A sample image is shown in fig. 3(a). The major preprocessing operations that have been done over each such image are discussed in the subsequent subsections.

### A. Median filtering

Median filter is a non-linear filter which replaces the gray value of a pixel by the median of the gray values of its neighbors. We have used 3 × 3 mask to get eight neighbors and their corresponding color values. As we have processed image three medians corresponding to red, green and blue colors are obtained and combined to get the color value of the replacing pixel. This operation removes salt-and-peeper noise from the image.

### B. Contrast enhancement

Contrast of each image is enhanced through histogram equalization technique, as discussed in [8]. We have applied the process for the intensity component of HSI color model of the whole image. This has the effect of improving the contrast of the image keeping the color information intact. The intensity level, within the range [0, 1], is divided equally in 100 levels having step size of 0.01. The probability of occurrence of intensity level $i_k$ can be written as:

$$p_i(i_k) = \frac{n_k}{n} \quad (2)$$

where k=0,1,2,….99 denotes the level number, $n_k$ is the number of pixels having intensity value $i_k$ and n is the total number of pixels in the image. The enhanced intensity value is:

$$I_k = \sum_{j=0}^{k} p_i(i_j) = \sum_{j=0}^{k} \frac{n_j}{n} \quad (3)$$

Since the histogram equalization sometimes introduce more intensity into or reduce intensity from the image the saturation level of the image is slightly increased after histogram equalization to get prominence in color [8]. The original image of fig. 3(a), when operated by the median filter, generates an intermediate image. After the enhancement of contrast of the said image fig. 3(b) is generated.

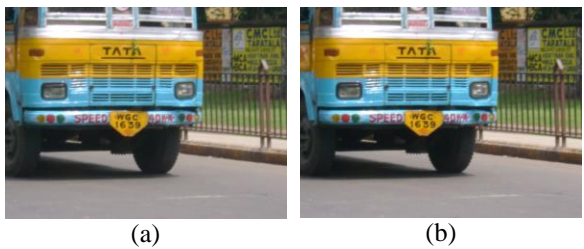

(a)          (b)

Fig. 3(a-b). A sample vehicle image and the corresponding result after preprocessing

*C. Preparation of license plate dataset*

After filtering and contrast improvement the images become ready to be dealt with. A large dataset containing the images of the commercial vehicles has been prepared. Now from each of the images the license plates have been extracted manually and stored as separate images. In this way a dataset of the license plates has been made. Fig. 4 shows a sample set of license plates of Indian commercial vehicles.

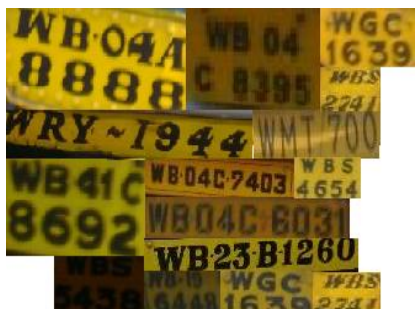

Fig. 4. License plates for preparing the dataset

*D. Preliminary statistical analysis over the license plates*

We have developed a technique for reading the license plates and computing the mean and standard deviation values for R, G, B, H, S and I components of each of the images, both for the yellow background and for the black characters separately. This is done by generating the frequency plot over the respective range scale for each of the color components and identifying the two highest peaks, generated for the dark characters and the yellowish background.

*E. Fixing the seed and tolerance values*

Once this statistical analysis has been done the next task is to identify the seed values for each color component and the allowable tolerance limits against each such seed values. The mean value of each component is used as seed value for that component for segmentation of the image. The standard deviation of each color component is used as tolerance limit against each such seed value for the respective color components.

*F. Image segmentation and connected component labeling*

Based on the seed values of the color components, as discussed in the previous subsection, the image is segmented into two classes: one indicating the license plate region and the other indicating no plate region. The basic classification depends on the values of H and S color components only. During the segmentation process, the pixels lying in the tolerance of H and S are labeled as $LABEL_{LP}$ indicating it as a part of license plate. The pixels lying outside the range of H and S is labeled as $LABEL_{NLP}$ indicating it as a part of no-plate region. After the image has been segmented, the connected components (in terms of pixels) are clustered. There may be one or more connected components depending on the segmentation result. These components are also of irregular shape. Now each of the connected components is out bounded by a rectangular box ($B_1$). As the boxes are out bounding the segments they may contain components of other segments as well. Thus the boxes may overlap each other.

*G. Finding the potential regions for the license plates*

To identify the potential region of interest for the license plates, the clusters of overlapping boxes need to be encompassed with still larger rectangular boxes ($B_2$). Each such box contains a set of connected boxes ($B_1$). Now there may be one or more $B_2$ boxes indicating the potential regions where the license plates may lie. To detect the correct regions of the license plates, following feature are considered:

i. Aspect ratio of the license plate
ii. Area of the license plate in terms of number of pixels it contains
iii. Fractional area of the license plate relative to the area of the whole image
iv. Average horizontal contrast density

All the above parameters are measured for each of the $B_2$ boxes to get the correct box that may contain the license plate. This box is marked as $B_3$. The following steps illustrate the algorithm for calculation of the average and maximum contrast of each $B_2$ box in the image frame.

*Steps*

National Conference on Computing and Communication Systems (COCOSYS-09)
CS10                                                                                                                             209```
Set maxcontrast = 0
For boxindex = 0 to MAX do
    contrast = 0
    For x = XMIN to XMAX do 16
    For y = YMIN to YMAX do 16
        [R₁, G₁, B₁, H₁, S₁, I₁] = RGB_HSI(x,y)
        label = Label(x,y)
        If label ≠ LABEL_LP then go to 16
        [R₂, G₂, B₂, H₂, S₂, I₂] = RGB_HSI(x+1,y)
        label = Label(x+1,y)
        If label ≠ LABEL_LP then go to 16
        dR = Abs(R₁-R₂), dG=Abs(G₁-G₂), dB=Abs(B₁-B₂)
        RGBSum = dR + dG + dB
        If RGBSum <= C_th then go to 16
        contrast = contrast + RGBSum
    Loop y
    Loop x
    BoxArea = (XMAX-XMIN)×(YMAX-YMIN)
    Avg_Contrast = contrast/BoxArea
    If Avg_Contrast > maxcontrast then maxcontrast =
        Avg_Contrast
Loop boxindex
```

As discussed above, within the segment having label as $LABEL_{LP}$ the contrast between two successive pixels is accounted. Here a new expression for calculation of contrast has been used. The sum of R, G and B components is calculated for two successive pixels and the difference between the two sums is considered as horizontal contrast. It is also considered that if the contrast is more than $C_{th}$ then only it is considered as a prominent edge and only those contrast are taken into account for calculating the average contrast of the box. This eliminates the rough and noisy variation of color even within the same segment. Large variation in color resulting high contrast will be generated by the digits and characters within the segments having label as $LABEL_{LP}$. Once the average and maximum contrast of each $B_2$ box is calculated, the next step is to detect the location of the correct $B_3$ box containing license plate. Aspect ratio, area and average horizontal contrast density features are considered for each of the $B_2$ boxes for successful localization of $B_3$ boxes. Values of different experimental thresholds used during localization of the license plate regions are discussed in the following section.

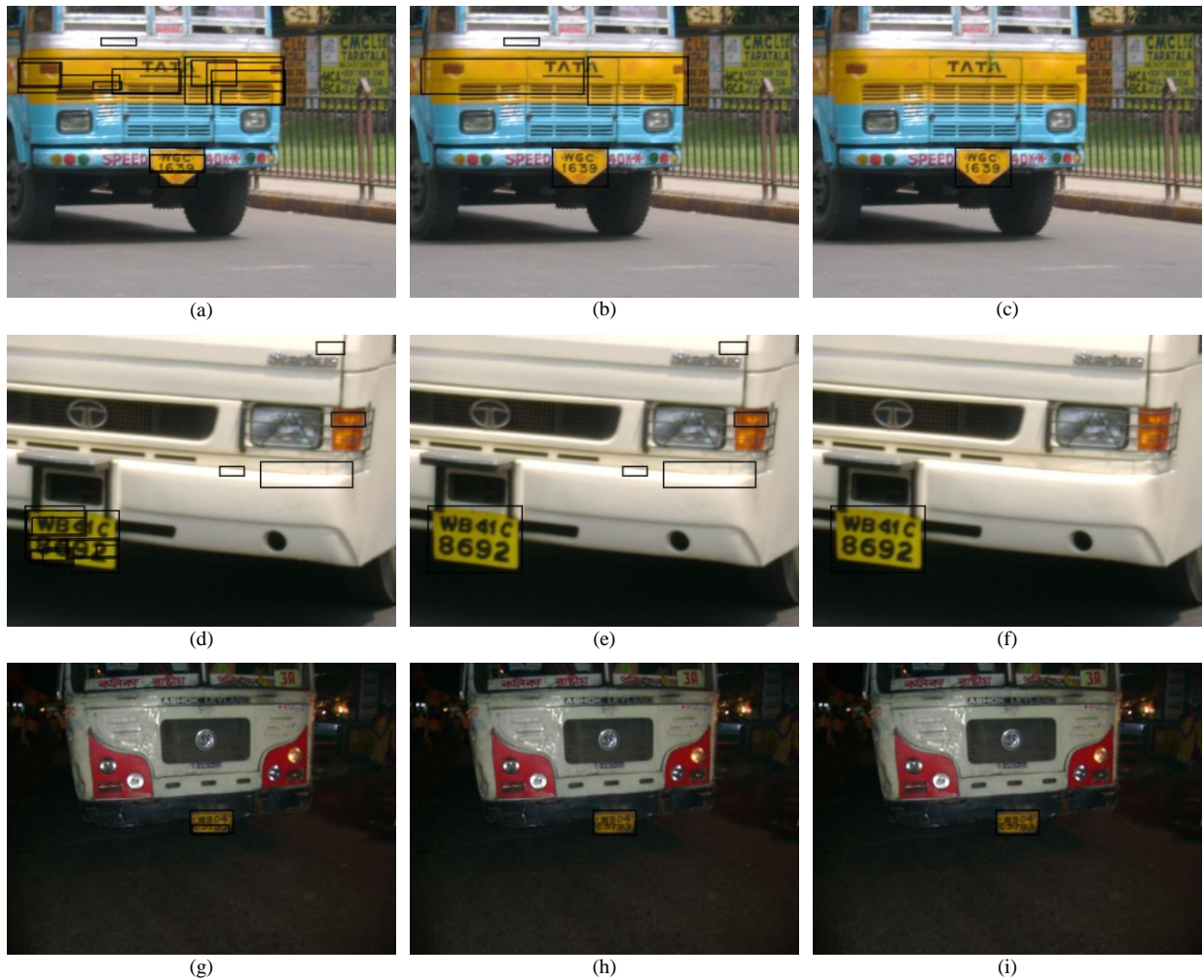

(a)  (b)  (c)
(d)  (e)  (f)
(g)  (h)  (i)

Fig. 5. Successfully localized license plate regions in vehicle images.
(a), (d), (g): vehicle images with 1st level of localization indicated by B1 boxes.



(b), (e), (h): vehicle images with 2nd level of localization indicated by B2 boxes.
(c), (f), (i): vehicle images with final localization of license plate regions indicated by B3 boxes.

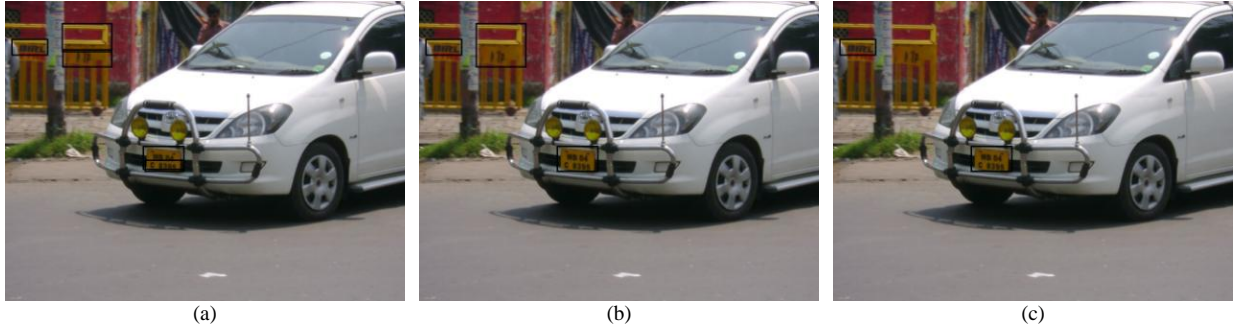

(a)          (b)          (c)

Fig. 6(a-c). Erroneously localized additional license plate regions by the B3 boxes.

III. EXPERIMENTAL RESULTS

For preparation of the dataset, vehicle images are captured through a fixed camera, installed by the road-side. True color bitmap images of resolution 840x630 pixels are captured with varying lighting conditions. The current image dataset consists of 500 such images capturing variety of vehicles. In the current experiment, we have excluded all the images of private vehicles from the image dataset.

Different experimental considerations for localization of potential license plate regions are estimated as follows:

*Minimum contrast of a potential $B_2$ box, $C_{th}$ = 100*
*Minimum area of a $B_3$ Box = 2000 pixels*
*Range of aspect ratio for $B_3$ boxes : 1.2 to 10*
*Maximum contrast difference in $B_2$ boxes, $C_{diff}$= 0.2*

Using the above considerations, experiments are conducted with the images of commercial vehicles from the generated dataset. Figures 5-6 shows different steps of experimental results on the original images, i.e., identification of $B_1$, $B_2$ and $B_3$ boxes. Fig. 5(a-i) shows some of the vehicle images with successfully localized license plate regions. The technique could successfully localize the license plate even in poor lighting conditions, as shown in fig. 5(g-i). Fig. 6(a-c) shows a sample image where an additional region is identified as a license plate. In many of the failure cases, the true license plate is identified along with some erroneous license plate regions. The technique also fails to localize the boundary of the license plates from some of the completely yellow colored vehicles for which license plate color is similar to the body color of such vehicles.

For computation of localization accuracy, we have considered inaccurate localization of license plates as the *false negative* cases. Identification of additional license plate regions, along with the true license plates, is identified as the *false positive* cases and finally the perfect localization of license plate regions as *true positive* cases. As observed from the experimentation on the collected dataset of vehicle images, the *false negative* rate is 10.6%, *false positive* rate is 16% and finally the *true positive* accuracy is 73.4%. If we ignore only the *false negative* cases, then the combined *positive* accuracy may be estimated as 89.4%, where the technique localizes the true license plates from vehicle images.

IV. CONCLUSION

In the current work, we have developed a simple and effective scheme for localization of license plate regions for Indian commercial vehicles. We have extensively applied the algorithm on 500 images. It is seen that in case of completely yellow colored vehicles the algorithm does not perform well because of the similarity of the color of the vehicle and the license plate, i.e., absence of prominent license plate boundaries. In some of such images the output is wrong localization, in some cases the plate region is detected as very large locality and in some of the cases other regions are falsely detected as license plate. Despite these difficulties, the technique fares well in most of the vehicle images, even in darker lighting conditions. Apart from the salt and peeper noise, as discussed in earlier sections, motion blur and other types of noises often degrade the image quality. Specific image enhancement algorithms may be employed in future to improve the overall performance of the developed system.

The technique can further be enhanced by incorporating shape based features for yellow/white colored license plates and employing soft computing techniques for automatic localization of license plate regions from commercial and private vehicle images. The localized license plate regions are to be subsequently processed by an effective *OCR* module for extraction of vehicle license numbers.

ACKNOWLEDGEMENT








## REFERENCES

[1] O. Martinsky, "Algorithmic and Mathematical Principles of Automatic Number Plate Recognition System", *B. Sc. Thesis*, BRNO University of Technology, 2007.
[2] Erik Bergenudd, "Low-Cost Real-Time License Plate Recognision for a Vehicle PC", *Master's Degree Project*, KTH Electrical Engineering, Sweden, December 2006.
[3] Cesar Garcia-Osorio, Jose-Francsico Diez-Pastor, J. J. Rodriguez, J. Maudes, "License Plate Number Recognition New Heuristics and a comparative study of classifier", cibrg.org/documents/Garcia08ICINCO.pdf.
[4] J. R. Parker and P. Federl, "An Approach to License Plate Recognition", Computer Science Technical Report(1996-591-1. I), 1996.
[5] H. Kawasnicka and B. Wawrzyniak, "License Plate Localization and Recognition in Camera Pictures", AI-METH 2002, Poland, November 2002.
[6] C. N. Anagnostopoulos, I. Anagnostopoulos, V. Loumos and E. Kayafas, "A license plate recognision algorithm for Intelligent Transport applications", www.aegean.gr/culturaltec/canagnostopoulos/cv/T-ITS-05-08-0095.pdf.
[7] http://www.htsol.com
[8] R. C. Gonzalez and R. E. Woods, Digital Image Processing, Pearson Education Asia, 2002.



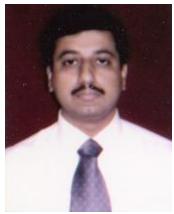
Satadal Saha received B. Sc. (Physics Hon's) from Bidhannagar College. Kolkata in 1995, completed B. Tech. in Applied Physics and M. Tech. in Optics and Optoelectronics from University of Calcutta in 1998 and 2000. In 2004, he joined as a Lecturer in the Department of Computer Science and Engg., MCKV Institute of Engg, Howrah and he is continuing his service there as an Assistant Professor. He has published a book titled Computer Network (New Delhi: Dhanpat Rai and Co. Ltd., 2008). His research areas of interest are image processing and pattern recognition. Mr. Saha is a member of IETE and CSI. He was also a member of the executive committee of IETE, Kolkata for the session 2006-08.

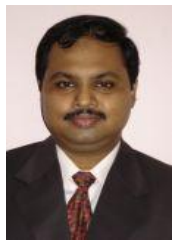
Subhadip Basu received his B.E. degree in Computer Science and Engineering from Kuvempu University, Karnataka, India, in 1999. He received his Ph.D. (Engg.) degree thereafter from Jadavpur University (J.U.) in 2006. He joined J.U. as a senior lecturer in 2006. His areas of current research interest are OCR of handwritten text, gesture recognition, real-time image processing.

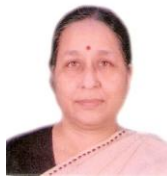
Mita Nasipuri received her B.E.Tel.E., M.E.Tel.E., and Ph.D. (Engg.) degrees from Jadavpur University, in 1979, 1981 and 1990, respectively. Prof. Nasipuri has been a faculty member of J.U since 1987. Her current research interest includes image processing, pattern recognition, and multimedia systems. She is a senior member of the IEEE, U.S.A., Fellow of I.E (India) and W.B.A.S.T, Kolkata, India.

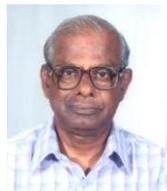
Dipak Kumar Basu received his B.E.Tel.E., M.E.Tel., and Ph.D. (Engg.) degrees from Jadavpur University, in 1964, 1966 and 1969 respectively. Prof. Basu has been a faculty member of J.U from 1968 to January 2008. He is presently an A.I.C.T.E. Emiretus Fellow at the CSE Department of J.U. His current fields of research interest include pattern recognition, image processing, and multimedia systems. He is a senior member of the IEEE, U.S.A., Fellow of I.E. (India) and W.B.A.S.T., Kolkata, India and a former Fellow, Alexander von Humboldt Foundation, Germany.